\documentclass[runningheads]{llncs}
\usepackage[T1]{fontenc}
\usepackage{graphicx}
\usepackage{booktabs}
\usepackage[misc]{ifsym}
\newcommand{\corr}{(\Letter)}
\usepackage{amsmath}
\usepackage{amssymb}
\usepackage{amsfonts}
\usepackage{caption}
\usepackage{subcaption}
\usepackage{booktabs}
\usepackage{graphicx}

\usepackage{researchpack}
\usepackage[square,numbers,comma,sort&compress]{natbib}
\usepackage{cleveref}
\usepackage{xcolor}
\usepackage{xspace}
\usepackage{algorithm}
\usepackage{algpseudocode}
\usepackage{bm}

\newcommand{\method}{\texttt{XIGL}\xspace}
\newcommand{\acronym}{eXplanatory Interactive Graph shortcut unLearning\xspace}

\newcommand{\GCN}{\texttt{GCN}\xspace}
\newcommand{\GIN}{\texttt{GIN}\xspace}
\newcommand{\SAGE}{\texttt{GraphSAGE}\xspace}
\newcommand{\GAT}{\texttt{GAT}\xspace}

\begin{document}

\title{Overcoming Shortcut Learning in Graph Neural Networks through Active Explanation Guidance}
\titlerunning{Overcoming Shortcuts in GNNs with Explanation Guidance}


\author{Taraneh Younesian\inst{1} \corr \and
Steve Azzolin\inst{2} \and
Antonio Longa\inst{3} \and
Francesco Ferrini\inst{2} \and
Vincenzo Marco De Luca\inst{2} \and
Stefano Teso\inst{2}}

\authorrunning{T. Younesian et al.}

\institute{
VU Amsterdam, Netherlands
\email{[t.younesian@vu.nl](mailto:t.younesian@vu.nl)}
\and
University of Trento, Italy
\email{{steve.azzolin,francesco.ferrini,vincenzomarco.deluca,stefano.teso}@unitn.it}
\and
UiT The Arctic University of Norway, Norway
\email{[antonio.longa@uit.no](mailto:antonio.longa@uit.no)}
}





\maketitle              

\begin{abstract}
Graph Neural Networks (GNNs) can solve prediction tasks by unintentionally exploiting \textit{shortcuts}---that is, edges, nodes, and features that correlate with but are not causal for the prediction---which compromise their reliability in out-of-distribution tasks.
We introduce \method (\acronym), an architecture-agnostic human-in-the-loop strategy for removing such shortcuts from GNNs.
Our key insight is twofold.  On the one hand, reliance on shortcuts can be detected by inspecting GNN explanations.  On the other hand, once made aware of such shortcuts, sufficiently expert users can provide tailored corrective feedback, which helps deconfound the model.
\method supports any query strategy; however, since corrective feedback can be expensive to acquire, we develop an active learning strategy for prioritizing explanations that are more likely to display shortcut behavior, lowering annotation and cognitive costs.
We showcase the effectiveness of \method, including both existing and proposed explanation-based strategies, on several GNN architectures. Our implementation is available online. \footnote{https://github.com/TYounesian/xilgraph.git}

\end{abstract}


\section{Introduction}

Graph Neural Network (GNN) classifiers can pick up on \textit{shortcuts}, also called \textit{confounders}, patterns that happen to correlate with the label in-distribution, allowing the model to achieve high accuracy but not generalize outside of it.  Reliance on such shortcuts---including watermarks \cite{lapuschkin2019unmasking}, metadata \cite{geirhos2020shortcut}, and simple input statistics \cite{pluska2024logical}---can prevent models from behaving properly upon deployment when the data distribution changes.
Since providing more (observational) training data is insufficient to resolve shortcuts, existing works address them by employing other kinds of data, such as examples from multiple domains \cite{arjovsky2019invariant, krueger2021outofdistribution}, that are not always available.

\begin{figure}[t]
    \centering
    \includegraphics[width=\linewidth]{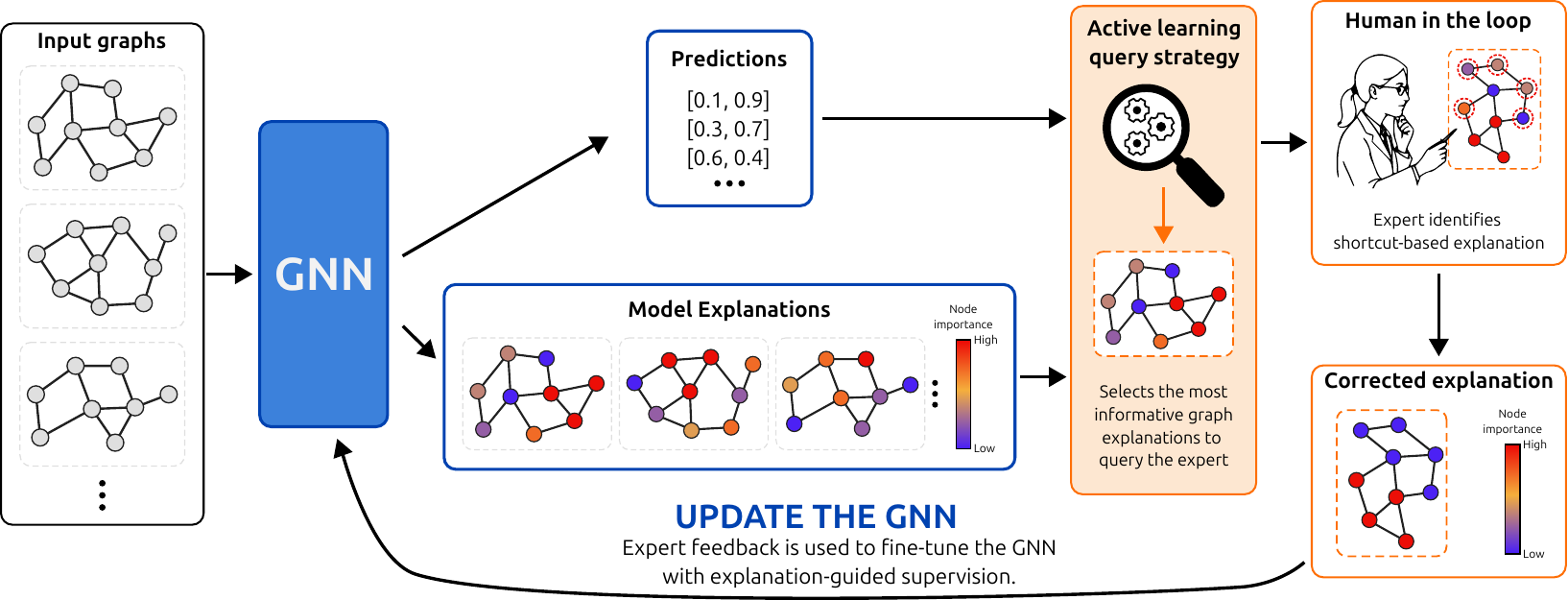}
    \caption{\textbf{Overview of \method}.  
        We first extract explanations from the GNN being trained over a small subset of the dataset; an active learning strategy then selects the most informative explanations for an expert to correct by indicating which nodes the model should not rely on and which nodes it should rely on instead; the corrections then fine-tune the GNN via an explanation-guided loss that removes shortcut dependence. The process iterates until the annotation budget is reached.
    }
    \label{fig:overview}
\end{figure}

We take a different route.
Specifically, we exploit the fact that, by unveiling the GNN's reasoning process, GNN \textit{explanations} can naturally expose its reliance on shortcuts \cite{azzolin2023global, geirhos2020shortcut, lapuschkin2019unmasking, pluska2024logical, ross2017right, de2025xai}.
We then design \method (\acronym), an active learning pipeline, illustrated in Figure \ref{fig:overview}, that iteratively
\textit{i}) employs an intuitive selection strategy to identify those instances whose explanations are more likely to reveal shortcut behavior;
\textit{ii}) requests a human annotator to correct these explanations by indicating what nodes of the input graph the GNN should \textit{not} rely on; and
\textit{iii}) adapts the GNN based on the user's corrections through an end-to-end differentiable loss function geared toward shortcut removal \citep{ross2017right, teso2023leveraging}.

Compared to existing work on explanation-based deconfounding of GNNs \cite{zhang2026quantifying}, which assumes that explanatory supervision is available for the entire training set, \method actively prioritizes annotating a subset of the data with those explanations that better capture dependency on shortcuts.  This avoids the need to annotate examples where the model is behaving properly, while focusing the annotator's effort on subgraphs for which feedback is needed and useful, thus reducing annotation and cognitive costs.

Focusing on graph classification tasks, while \method supports any query strategy, our experiments demonstrate how \method's explanation-based query strategies mostly improve on existing active learning baselines, which select instances based purely on prediction-based informativeness across several GNN architectures.



\section{Preliminaries}
\label{sec:preliminaries}

Let $G = (\mathcal{V}, \mathcal{E},X)$ be an (undirected) \textit{graph}, where $\mathcal{V}=\{v_1,...,v_N\}$ is the set of nodes, $\mathcal{E} \subseteq \mathcal{V} \times \mathcal{V}$ is the set of edges, $X \in \mathbb{R}^{N \times d}$ is the node features, and $y \in \{1,\ldots,l\}$ is the label assigned to $G$.
We tackle \textit{graph classification} problems where the goal is to learn a function $f: \mathcal{G \to \mathcal{Y}}$ mapping from the set of possible graphs $\mathcal{G}$ to the set of labels $\mathcal{Y}$ from training examples of annotated graphs.


We focus on scenarios in which the input graphs $G$ encode both ``causal'' and ``spurious'' subgraphs, denoted $C$ and $S$, respectively:  whereas $C$ determines the ground-truth label $y$, $S$ only correlates with it.
To build intuition, consider \texttt{CPatchMNIST} \cite{azzolin2026gnn}: here, the graphs $G$ represent handwritten digits in terms of superpixels (nodes) and their adjacency relation (edges).  Node features encode the average color of the corresponding superpixel, and the label $y$ is the digit itself.  Crucially, the graphs are manipulated such that the color of the rightmost superpixels correlates with the digit, and as such act as shortcuts.

%
%
%
Neural nets are known to be biased to learn simpler solutions \cite{yang2024identifying}. In real-world settings, the confounder $S$ is often easier to learn than the causal one $S$.  In our example, since the spurious superpixels are highly discriminative, GNNs tend to exploit them for prediction.  This yields excellent in-distribution performance but near-random accuracy in unseen test instances where color is removed.

Throughout, we refer to the nodes in the causal subgraph $C$ as \textit{relevant} nodes and denote them $\calV_{r}$, and to the nodes of its complement as \textit{irrelevant} nodes, denoting them $\mathcal{V}_{ir}$.  The latter covers the spurious subgraph $S$, as well as the rest of the graph.
Ideally, we wish our GNNs to make predictions for the ``right reason'', that is, by relying on the causal subgraph rather than the spurious one.


\section{Deconfounding through Active Explanation}
\label{sec:method}

We aim to deconfound the model by guiding it to
not use shortcuts. We do so by penalizing the impact of irrelevant subgraphs of nodes $V_{ir}$ on the model. 
As an indication of the impact of nodes, we use input gradients \cite{Pope_2019_CVPR, zhang2026quantifying}, which we adopt among the many possible GNN explainers (see \cite{longa2025explaining, survey} for a survey) for their simplicity and because they do not require training an ad-hoc explainer. We then minimize this value for irrelevant nodes in each graph, as follows:
\begin{equation}
    \begin{aligned}
    \mathcal{L} &= \mathcal{L}_{ce} + \lambda\mathcal{L}_e   = -\sum_{i=1}^{l} \hat{y}_{i} \log(\hat{y}_{i}) + \lambda \frac{\sum_{j=1}^{N}(1-e_j)s_j^2}{\sum_{j=1}^{N}s_j^2},
    \end{aligned}
    \label{lossfunc}
\end{equation}
where $\hat{y}_i$ is the predicted probability of class $i$, $s_j=\sum_{i=1}^{l}\nabla_{j} \log(\hat{y}_i)$ is sum of the input gradients, $\lambda$ is a scaling term to balance between classification and explanation loss, and $\mathbf{e}\in \{0,1\}^N$ is the ground-truth explanation of the graph $G$. In particular, for each node $j$, $\mathbf{e}$ indicates if $j$ belongs to the relevant or irrelevant subgraph, i.e., $e_j$ is $1$ if $v_j$ is relevant, i.e., $v_j \in \mathcal{V}_r$, and is $0$ otherwise. In words,
minimizing $\mathcal{L}_e$ corresponds to minimizing the ratio of the input gradients of irrelevant nodes and the total input gradients across all nodes. In contrast to \cite{zhang2026quantifying}, we incorporate a normalization term in the denominator to prevent the trivial solution in which the input gradients of all nodes, including the relevant ones, collapse to zero. The loss function above aims to balance learning the labels with learning the truly relevant subgraph. Computing the total loss gradient during backpropagation yields second-order gradients with respect to the input. In practice, the resulting computational overhead was small.

\subsection{Passive vs. Active Explanation Supervision}
\label{al}

So far, we assumed access to the ground-truth explanation $\mathbf{e}$ for every graph in the training set. We call this scenario \emph{passive explanation supervision}, since our model passively uses all instances to learn from. Obtaining ground-truth explanations for every instance is costly and time-intensive, as it requires expert annotation to determine the task-relevant nodes in each graph. Therefore, this assumption is impractical in real-world scenarios. 

To address this limitation, we introduce an annotation budget $B$, which restricts the total number of graphs for which ground-truth explanations\footnote{We use \emph{ground-truth explanation} for simplicity. In practice, the explanation corrections provided by users do not need to match the true causal mechanism in the data, as they may not know the actual cause of the phenomenon being modeled. Yet, they may reliably indicate which nodes the
model \emph{should} rely on to avoid clear shortcuts.} can be collected.
Instead of annotating the entire dataset, we select a small, informative subset of graphs so that training on this subset yields a highly accurate model that captures the underlying causal patterns.

We employ \emph{active learning} (AL) \cite{settles2009active, li2024survey, song2023no, zhang2022galaxy} to identify informative instances. While \method is agnostic to the choice of query strategy and can be combined with any active learning method, we introduce two novel explanation-based query strategies that leverage explanation uncertainty to identify instances whose explanation corrections are expected to provide the greatest benefit to the model. We include two standard query strategies in \method for comparison. 

Let $\mathcal{D}_U=\{(G_i,y_i)\}_{i=1}^{N}$ be the dataset of graphs and their labels without their ground-truth explanation. The process begins by training the model on an initial small set of $q$ graphs for which we have the explanations denoted as $\mathcal{D}_{\mathrm{exp}}=\{(G_i,y_i,E_i)\}_{i=1}^{q}$. Then, among the remaining rest of $\mathcal{D}_U$, we select a batch of the most informative graphs and query an expert for their ground-truth explanations. The newly annotated batch is added to the previously annotated set $\mathcal{D}_{\mathrm{exp}}$, and the model is subsequently fine-tuned on this set using Eq. \ref{lossfunc}. We repeat this procedure for $T \le B$ iterations, ensuring that the total number of annotated instances does not exceed the budget. Algorithm \ref{alg:eg-al} shows the steps of \method's active learning. We investigate two classes of query strategies: prediction-based and explanation-based. The former are established active learning methods, whereas the latter are novel strategies proposed in this paper. The details of each strategy are provided below:

\begin{algorithm}[t]
\caption{Active Learning Steps in \method}
\label{alg:eg-al}
\begin{algorithmic}[1]
\Require Labeled graph dataset $\mathcal{D}_U=\{(G_i,y_i)\}_{i=1}^{N}$, initial query budget $q$, query budget $B$, active learning iteration $T$, query strategy $\mathcal{S}$
\Ensure Updated GNN model $f$ 

\State Select $q$ graphs and obtain their explanation correction:
\[
\mathcal{D}_{\mathrm{exp}}
=
\{(G_i,y_i,E_i)\}_{i=1}^{q}.
\]

\State Train an initial GNN $f_\theta$ using: $\mathcal{L}=\mathcal{L}_{\mathrm{ce}}+\lambda \mathcal{L}_{\mathrm{e}}$.

\For{each iteration $t=1,\dots,T$}
    \State Generate predictions and explanations for graphs in
    $\mathcal{D}_U\setminus\mathcal{D}_{\mathrm{exp}}$.
    
    \State Select a query set $\mathcal{Q}$ of size $B/T$ according to $\mathcal{S}$ (Section \ref{al}).
    
    \State Obtain explanation correction for graphs in $\mathcal{Q}$.
    
    \State Update the explanation-supervised set:
    \[
    \mathcal{D}_{\mathrm{exp}}
    \leftarrow
    \mathcal{D}_{\mathrm{exp}}
    \cup
    \{(G_i,y_i,E_i):G_i \in \mathcal{Q}\}.
    \]
    
    \State Fine-tune the GNN using $\mathcal{L}=\mathcal{L}_{\mathrm{ce}}+\lambda \mathcal{L}_{\mathrm{e}}$.
\EndFor
\end{algorithmic}
\end{algorithm}

\subsubsection{Prediction-based strategies}
\begin{itemize}
    \item \textbf{Maximum Classification Entropy} (MaCE) 
    is one of the most popular uncertainty-based query strategies. MaCE queries graphs that the GNN is highly uncertain to classify, using the Shannon entropy: 
\begin{equation}
G^* = \argmax_{G\in\mathcal{D}_U} - \sum_{i=1}^l p(y_i \mid G) \log p(y_i \mid G)
\end{equation}
 \item \textbf{Random Sampling} 
 randomly queries graphs according to a uniform distribution. Although random sampling does not make use of predictive information, we include it in this category as a baseline for comparison.
\end{itemize}

\subsubsection{Explanation-based strategies} 

\begin{itemize}
    \item \textbf{Maximum Explanation Entropy} (MaEE)
    queries the graphs that the GNN is most uncertain in its \textit{explanation}. Specifically, it prioritizes graphs in which input gradients 
    are most uniform across nodes, as measured by the entropy of softmax-normalized node-wise input gradients:
\begin{equation}
G^* = \argmax_{G\in\mathcal{D}_U} - \frac{\sum_{i=1}^{|\mathcal{V}|} p_i\log p_i}{\log |G|},
\end{equation}
where $p_i=\frac{\exp(s_i)}{\sum{j\in G}\exp(s_j)}$ is the softmax-normalized input gradient $s_i$.  Since the number of nodes in each graph varies, we normalize the entropy to compare across graphs.

    \item \textbf{Minimum Explanation Entropy} (MiEE)
    queries the graphs that the GNN is most \textit{certain} in its explanation. We adopt this strategy based on the simplicity bias of neural networks \cite{yang2024identifying}, which causes models to preferentially learn simpler yet spurious patterns. As a result, the model may quickly overfit to confounders and produce overly confident explanations that incorrectly attribute the prediction to the confounding features.
\begin{equation}
G^* = \argmin_{G\in\mathcal{D}_U} - \frac{\sum_{i=1}^{|\mathcal{V}|} p_i\log p_i}{\log |G|}.
\end{equation}

\end{itemize}


\section{Experiments}
\label{sec:experiments}

\begin{figure}[t]
  \centering

  \begin{subfigure}[b]{0.45\textwidth}
    \centering
    \includegraphics[width=\textwidth]{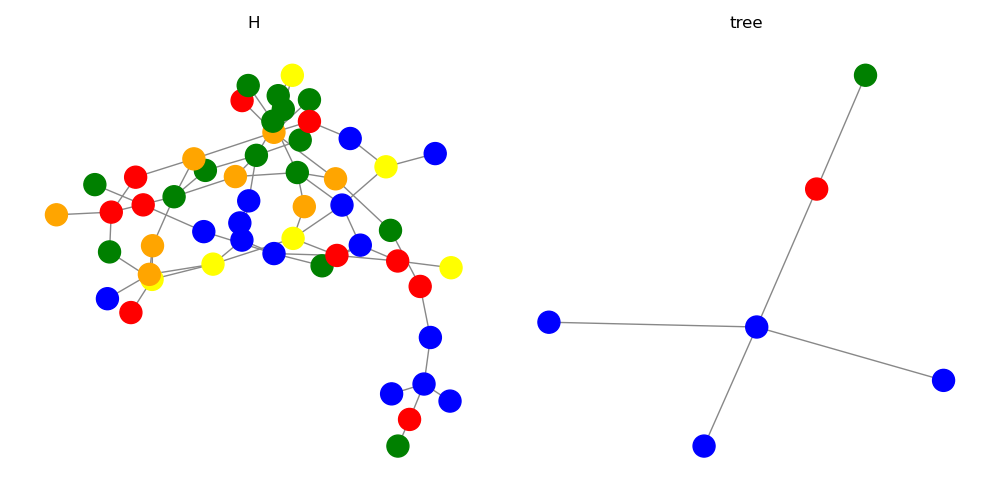}
    \caption{No confounder, $y=0$}
    \label{fig:no-conf-y0}
  \end{subfigure}
  \hfill
  \begin{subfigure}[b]{0.45\textwidth}
    \centering
    \includegraphics[width=\textwidth]{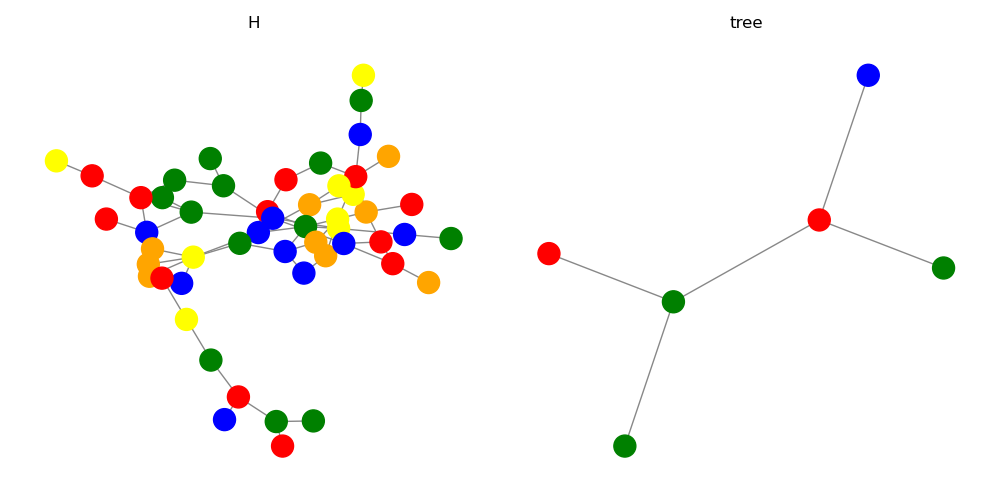}
    \caption{No confounder, $y=1$}
    \label{fig:no-conf-y1}
  \end{subfigure}

  \vspace{0.5em}

  \begin{subfigure}[b]{0.45\textwidth}
    \centering
    \includegraphics[width=\textwidth]{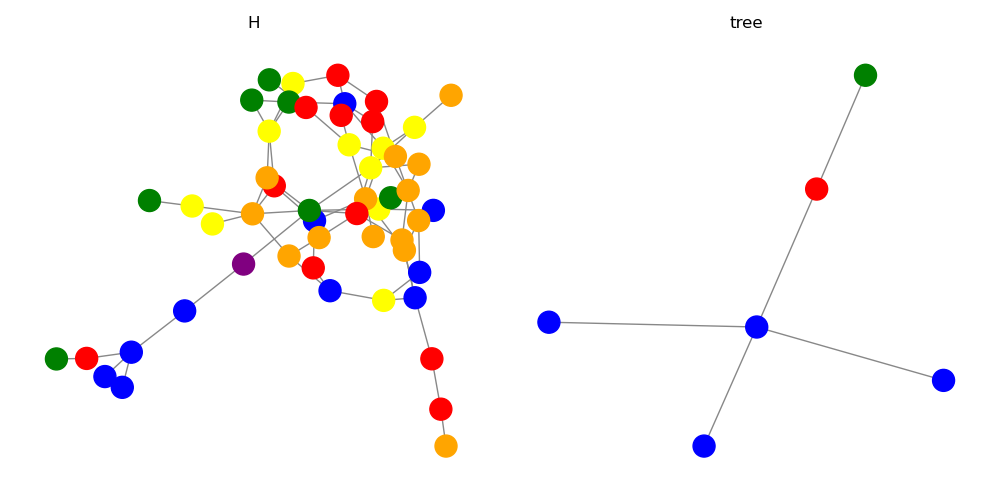}
    \caption{Confounder with a new color, $y=0$}
    \label{fig:conf-new-color-y0}
  \end{subfigure}
  \hfill
  \begin{subfigure}[b]{0.45\textwidth}
    \centering
    \includegraphics[width=\textwidth]{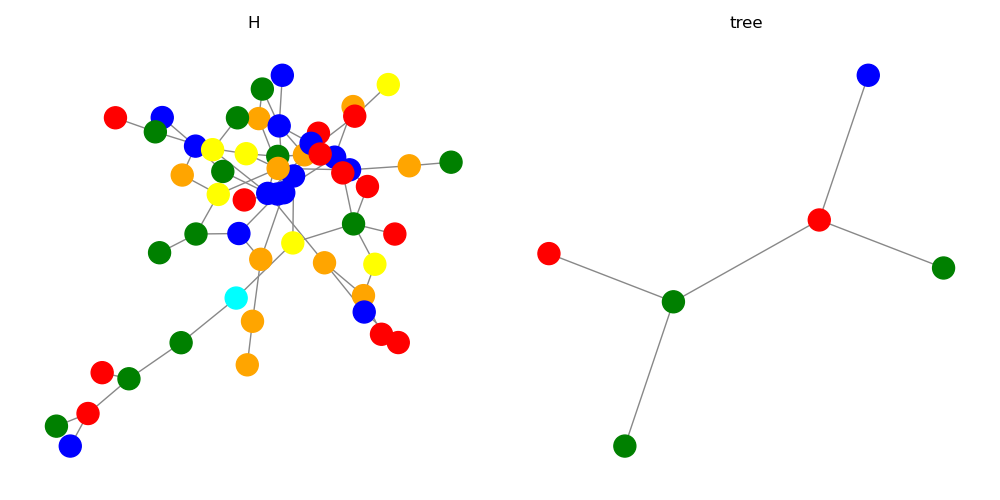}
    \caption{Confounder with a new color, $y=1$}
    \label{fig:conf-new-color-y1}
  \end{subfigure}

  \caption{\texttt{ER-color} dataset examples with and without confounders. The trees shown are the ground-truth explanations for each class. As confounders, for $y=0$, a purple node is added to the graph, and for $y=1$, a cyan node is added.}
  \label{fig:conf}
\end{figure}

\textbf{Datasets}  We use one synthetic and one real-world dataset for our experiments:
%
\underline{\texttt{ER-color}}
is a synthetic dataset 
comprising $1000$  Erdős-Rényi graphs with $50$ nodes and edge probability of $0.05$. We randomly assign the colors red, blue, green, yellow, and orange to the nodes using a uniform distribution. We split the dataset into $70\%$, $15\%$, and $15\%$ for train, validation, and test sets, respectively. We then randomly attach two motifs to the base graphs, each representing a label. 
Hence, these motifs represent the \textit{right reasons}.
To create confounders that induce a distribution shift between the train and validation/test sets, we add $j$ purple nodes for label $0$ and $j$ cyan nodes for label $1$ as confounders only in the train set, while keeping the validation and test sets unchanged. Node features are represented as one-hot encoding of colors. For confounding colors, the active feature value is set to $100$ rather than $1$, thereby amplifying the confounding signal. We use $j=1$ for passive learning experiments, and for active learning, we randomly select $j$ where $j \sim \textsf{Uniform}({1, \ldots ,25})$. The reason for the varying number of confounders in active learning experiments is to create different levels of informativeness among training instances. Figure \ref{fig:conf} shows the cases with and without confounders, i.e., training or validation/test sets, respectively.

\underline{\texttt{CPatchMNIST}}
\cite{azzolin2026gnn} is an extension of \texttt{MNISTsp} \cite{knyazev2019understanding}, itself
a conversion of the MNIST dataset to graphs via a superpixelation algorithm. In \texttt{CPatchMNIST}, on the other hand, the top left and bottom right superpixels are colored according to the graph label in the training set to represent confounders. However, in validation and test sets, these colors are randomized.

\noindent \textbf{Experimental Setup} We examined four GNN architectures—\GCN \cite{kipf2017semisupervised}, \GIN \cite{xu2018how}, \emergencystretch=2em \SAGE \cite{hamilton2017inductive}, and \GAT \cite{velickovic2018graph}—to assess both their vulnerability to shortcut learning and their ability to recover from it by avoiding spurious dependencies through explanation guidance. 
Further implementation details are available in the Appendix. For the active learning experiments, we query $5$ instances per round over $20$ rounds for \texttt{ER-color}, and $50$ instances per round over $10$ rounds for \texttt{CPatchMNIST}. We observed that when the annotation budget is large, the gap between different query strategies largely vanishes. Intuitively, this occurs because the informative instances selected by active learning methods constitute a subset of the data that is increasingly likely to be included in larger randomly sampled subsets. As a result, the benefits of querying informative instances become less noticeable.

\begin{table}[!t]
    \caption{\textbf{Effect of passive supervision in \method across datasets.}
    Accuracy (\%) on the unconfounded test split. Results are averaged with standard deviation across 5 seeds.}
    \label{tab:passive-learning}
    \centering
    \setlength{\tabcolsep}{4pt}
    \begin{tabular}{l|ll|ll}
        \toprule
        & \multicolumn{2}{c|}{\texttt{ER-color}}
        & \multicolumn{2}{c}{\texttt{CPatchMNIST}} \\
        \cmidrule(lr){2-3}\cmidrule(lr){4-5}
        & {\bf No Sup.} & {\bf Passive}
        & {\bf No Sup.} & {\bf Passive} \\
        \midrule
        \GCN  & $60.40{\scriptstyle \pm 11.36}$ & $\bm{71.20} {\scriptstyle \pm 20.73}$
              & $26.28 {\scriptstyle \pm 12.58}$ & $ \bm{30.57} {\scriptstyle \pm 2.90}$ \\
        \GIN  & $52.93 {\scriptstyle \pm  4.23}$  & $\bm{74.20} {\scriptstyle \pm 21.18}$
              & $0.20 {\scriptstyle \pm 0.12}$ & $\bm{33.38} {\scriptstyle \pm 1.13}$ \\
        \SAGE & $56.80 {\scriptstyle \pm 14.24}$ & $\bm{70.53} {\scriptstyle \pm 17.76}$
              & $\bm{31.5} {\scriptstyle \pm 6.85}$  & $30.68 {\scriptstyle \pm 1.88}$ \\
        \GAT  & $50.93 {\scriptstyle \pm 4.21}$  & $\bm{74.27} {\scriptstyle \pm 18.74}$
              & $22.61 {\scriptstyle \pm 3.93}$ & $\bm{25.25} {\scriptstyle \pm9.38}$ \\
        \bottomrule
    \end{tabular}
\end{table}


\subsection{Results}

\subsubsection{Passive Explanation Supervision}
Table \ref{tab:passive-learning} shows the test accuracy for \texttt{ER-color} and \texttt{CPatchMNIST} for different setups and GNN architectures, respectively. As the results show, passive supervision generally improves the accuracy compared to no-explanation supervision (No Sup.) and their ability to generalize to unconfounded data. We noticed that passive supervision's performance improves further if we first train the models for a few epochs only on $\mathcal{L}_e$, followed by training on $\mathcal{L}_{ce}+\lambda\mathcal{L}_e$, refer to Table \ref{passive-hyper} in Appendix \ref{app1} for more details. The results shown in Table \ref{tab:passive-learning} under Passive correspond to this setup. As the results show, on \texttt{ER-color}, most models perform near the chance level without explanation supervision but improve substantially with explanation supervision. This indicates that the models primarily rely on shortcut information and largely ignore the causal patterns, whereas explanation supervision guides them toward the causally relevant features.

On \texttt{CPatchMNIST}, most models perform slightly above chance level even without supervision, indicating that the causal signal can be learned to a limited degree. However, explanation supervision clearly improves performance for all architectures except for \SAGE. This effect is particularly evident for \GIN, which quickly overfits to the shortcuts and suffers a sharp drop in test accuracy, while explanation supervision in \method helps it recover and generalize better. While early stopping can reduce overfitting, it also limits the extent to which reliance on shortcuts becomes apparent. To expose this behavior and evaluate the robustness of explanation supervision, we do not use early stopping.

\subsubsection{Active Explanation Supervision}
Table \ref{tab:active-learning} shows the results of the different query strategies across both datasets and all GNNs. As the results indicate, the effectiveness of the query strategies varies across datasets and models. Nevertheless, active query selection generally outperforms random sampling in most settings. On \texttt{CPatchMNIST}, the explanation-based strategies consistently outperform the prediction-based strategies across all GNNs. Interestingly, random sampling achieves the best performance for \GIN in both datasets. We leave a systematic investigation of this effect for future work. Overall, MiEE emerges as the most consistently high-performing query strategy across models and datasets. These results highlight the potential of explanation-based active learning and motivate further investigation into explanation-driven query strategies, particularly those targeting highly confident yet potentially incorrect explanations.

As expected, the active learning component in \method generally achieves lower performance than passive learning due to its substantially smaller annotation budget. Nevertheless, the gap is relatively small, despite active learning using less than $20\%$ of the training data. These results suggest that active learning can achieve performance comparable to passive learning while requiring significantly fewer annotated explanations.




\begin{table}[!t]
    \caption{\textbf{Effect of AL strategies of \method across datasets.}
    Accuracy (\%) on the unconfounded test split. Results are averaged with standard deviation across 5 seeds.}
    \label{tab:active-learning}
    \centering
    \resizebox{\columnwidth}{!}{%
    \begin{tabular}{l|llll|llll}
        \toprule
        & \multicolumn{4}{c|}{\texttt{ER-color}}
        & \multicolumn{4}{c}{\texttt{CPatchMNIST}} \\
        \cmidrule(lr){2-5}\cmidrule(lr){6-9}
        & {\bf Random} & {\bf MaCE} & {\bf MaEE} & {\bf MiEE}
        & {\bf Random} & {\bf MaCE} & {\bf MaEE} & {\bf MiEE} \\
        \midrule
        \GCN  & $58.13 {\scriptstyle \pm 21.07}$ & $54.67 {\scriptstyle \pm 9.98}$
              & $59.87 {\scriptstyle \pm 20.43}$ & $\bm{64.00} {\scriptstyle \pm 13.93}$
              & $26.52 {\scriptstyle \pm 3.48}$ & $22.56 {\scriptstyle \pm 6.21}$ & $28.21 {\scriptstyle \pm 2.10}$ & $\bm{28.75} {\scriptstyle \pm 0.73}$ \\
        \GIN  & $\bm{68.80} {\scriptstyle \pm 22.93}$ & $63.87 {\scriptstyle \pm 12.40}$
              & $66.67 {\scriptstyle \pm 27.13}$ & $66.93 {\scriptstyle \pm 19.50}$
              & $\bm{32.74} {\scriptstyle \pm 2.80}$ & $25.50 {\scriptstyle \pm 3.88}$ & $30.87 {\scriptstyle \pm 2.61}$ & $27.77 {\scriptstyle \pm 2.77}$ \\
        \SAGE & $59.87 {\scriptstyle \pm 16.67}$ & $\bm{73.20} {\scriptstyle \pm 23.74}$
              & $55.20 {\scriptstyle \pm 7.49}$ & $56.27 {\scriptstyle \pm 11.90}$
              & $25.86 {\scriptstyle \pm 2.30}$ & $22.57 {\scriptstyle \pm 8.46}$ & $26.38 {\scriptstyle \pm 5.25}$ & $\bm{28.28} {\scriptstyle \pm 3.94}$ \\
        \GAT  & $52.67 {\scriptstyle \pm 14.12}$ & $\bm{60.40} {\scriptstyle \pm 22.10}$
              & $53.87 {\scriptstyle \pm 9.81}$ & $50.13 {\scriptstyle \pm 1.79}$
              & $26.90 {\scriptstyle \pm 2.34}$ & $26.59 {\scriptstyle \pm 1.22}$ & $\bm{29.20} {\scriptstyle \pm 11.64}$ & $28.57 {\scriptstyle \pm 5.86}$ \\
        \bottomrule
    \end{tabular}
    }
\end{table}

\section{Discussion, Related Work, \& Conclusion}
\label{sec:conclusion}

Our work suggests that, by exploiting expert corrections to model explanations, \method can help remove shortcut dependencies from models, and that it can reduce annotation costs compared to passive baselines.
In future work, we plan to extend the experiments to more datasets and GNN architectures, and specifically to self-explainable GNNs \citep{miao2022interpretable, tai2025redundancy, tai2026selfconsistency}.
Moreover, we plan to study the impact of explanation faithfulness \citep{Pope_2019_CVPR} on debiasing success: if an explanation does not capture the model's actual reasoning, then corrections to it may not help improve the model's behavior \citep{azzolin2025reconsidering}.  This is relevant for both post-hoc explainers \citep{adequately} and self-explainable architectures \citep{azzolin2026gnn}. Additionally, evaluating different kinds of shortcuts and real-world datasets is planned for future work.


\begin{credits}
\subsubsection{\ackname}
Taraneh Younesian was funded by Huawei DREAMS Lab. All content represents the opinion of the authors, which is not necessarily shared nor endorsed by their respective employers and/or sponsors in Huawei DREAMS Lab. Antonio Longa was supported by the Research Council of Norway through its Centre of Excellence Integreat - The Norwegian Centre for knowledge-driven machine learning, project number 33264. Funded by the European Union. Views and opinions expressed are however those of the author(s) only and do not necessarily reflect those of the European Union or the European Health and Digital Executive Agency (HaDEA). Neither the European Union nor the granting authority can be held responsible for them. Grant Agreement no. 101120763 - TANGO
\subsubsection{\discintname}
%
The authors have no competing interests to declare that are
relevant to the content of this article.
\end{credits}

%
%
\bibliographystyle{splncs04}
\bibliography{mybibliography,explanatory-supervision}
%

\newpage
\appendix
\section{GNN Details} \label{app1}
In this section, we present the hyperparameters for different GNN architectures across both datasets in the passive supervision scenario. Table \ref{passive-hyper} shows these values. We set the batch size to $16$ and $256$ for \texttt{ER-color} and \texttt{CPatchMNIST}, respectively, across all models and models. We evaluate all methods in a wide range of learning rates: $(1e-7, 1e-3)$ and $\lambda$ between $1$ and $1000$. For a fair comparison, we fix the number of epochs for the passive methods and set a fixed value across the active query strategies. As mentioned before, because early stopping can halt training before the effect of shortcuts becomes apparent, we deliberately do not use it. 
We plan to release the full codebase to reproduce our experiments upon acceptance.

\begin{table*}[t]
\centering
\caption{Hyperparameter settings for different GNN architectures for the passive supervision setup on \texttt{ER-color} and \texttt{CPatchMNIST}. ``H'' indicated the hidden dimension, ``Aggr.'' indicated the aggregation function of the GNN, ``Start E.'' indicates the starting epoch for training with both losses, and ``E''. indicated the total training epochs. }
\label{tab:gnn_hyperparams}
\resizebox{\textwidth}{!}{%
\begin{tabular}{lccccccccccccccccc}
\toprule
& \multicolumn{8}{c}{\texttt{ER-color}} & \multicolumn{8}{c}{\texttt{CPatchMNIST}} \\
\cmidrule(lr){2-9} \cmidrule(lr){10-16}
 &
\# Layers & H & Dropout & Aggr. & LR & $\lambda$ & Start E. & \# E &
\# Layers & H & Dropout & Aggr. & LR & $\lambda$ & Start E. & \# E\\
\midrule
GCN  & 3& 100& 0.3& add & $9e-7$& 10& 100 & 400& 3& 128& 0.3& add & $3e-6$&$19.22$ & 30 &500\\
GIN  & 3& 64 & 0  & mean&$5e-4$ & 10& 50 & 300 & 5& 256& 0  & mean& $8e-7$& $38.77$& 30&300\\
SAGE & 3& 128& 0  & add & $6e-6$& 1& 30 & 300 & 5& 128& 0  & add & $3e-6$& $71.77$& 70& 500\\
GAT  & 3& 128& 0  & add & $3.e-5$&1 & 0 & 300 & 3& 64 & 0  & mean& $5e-5$& $22.56$& 0&200\\
\bottomrule
\end{tabular}%
}
\label{passive-hyper}
\end{table*}

\section{AL Details}
For active learning experiments, we set $q$ to $10$ for both datasets. For \texttt{ER-color}, we query 5 instances per round for 20 rounds and 30 epochs per round. For \texttt{CPatchMNIST}, we query 50 instances per round for 10 rounds and 30 epochs per round. These settings hold for all models. The learning rates and $\lambda$ (varying in the same range as the passive experiments) for each GNN, query strategy, and dataset combination are shown in Table \ref{tab:al_hyperparams}.

\begin{table*}[t]
\centering
\caption{Training hyperparameters for different GNN architectures on \texttt{ER-color} and \texttt{CPatchMNIST}.}
\label{tab:al_hyperparams}
\resizebox{\textwidth}{!}{%
\begin{tabular}{lcccccccccccccccc}
\toprule
& \multicolumn{8}{c}{\texttt{ER-color}} &
  \multicolumn{8}{c}{\texttt{CPatchMNIST}} \\
\cmidrule(lr){2-9} \cmidrule(lr){10-17}
& \multicolumn{2}{c}{Random}
& \multicolumn{2}{c}{MaCE}
& \multicolumn{2}{c}{MaEE}
& \multicolumn{2}{c}{MiEE}
& \multicolumn{2}{c}{Random}
& \multicolumn{2}{c}{MaCE}
& \multicolumn{2}{c}{MaEE}
& \multicolumn{2}{c}{MiEE} \\
\cmidrule(lr){2-3}\cmidrule(lr){4-5}\cmidrule(lr){6-7}\cmidrule(lr){8-9}
\cmidrule(lr){10-11}\cmidrule(lr){12-13}\cmidrule(lr){14-15}\cmidrule(lr){16-17}

& LR & $\lambda$
& LR & $\lambda$
& LR & $\lambda$
& LR & $\lambda$
& LR & $\lambda$
& LR & $\lambda$
& LR & $\lambda$
& LR & $\lambda$ \\
\midrule
GCN  &$1e-5$ & $1.09$&$3e-5$&$1.19$ &$7e-6$ &$26.40$ &$5e-5$ &$25.21$ & $6e-7$& $251.75$& $6e-6$&$614.48$&$6e-7$ &$317.17$ & $8e-7$&$446.70$ \\
GIN  &$2e-4$ &$29.91$&$7e-4$ &$1.06$ & $6e-4$&$81.08$ &$2e-4$ &$2.51$ &$6e-7$&$162.74$& $2e-6$ & $672.56$& $3e-6$&$606.85$& $6e-7$& $212.88$ \\
SAGE & $5e-5$ &$23.91$ &$1e-4$ &$20.10$ &$5e-5$ &$22.23$ &$8e-5$ & $57.29$&$9e-6$ &$325.70$ & $3e-5$&$380.74$ & $2e-5$ & $850.60$& $3e-6$& $102.23$  \\
GAT  &$9e-5$ &$751.34$ & $6e-5$& $565.93$& $5e-5$& $565.52$ & $9e-6$ & $247.76$ & $4e-5$& $929.88$& $2e-5$&$130.81$ &$7e-5$ &$482.22$ &$1e-4$ & $961.77$\\
\bottomrule
\end{tabular}%
}
\end{table*}

\end{document}